\pdfoutput=0 

\documentclass[conference]{IEEEtran}
\IEEEoverridecommandlockouts

\usepackage{cite}
\usepackage{amsmath,amssymb,amsfonts}
\usepackage{algorithmic}
\usepackage[dvips]{graphicx}          
\usepackage{textcomp}
\usepackage{xcolor}
\PassOptionsToPackage{hyphens}{url}
\usepackage[dvips,hidelinks]{hyperref} 
\def\BibTeX{{\rm B\kern-.05em{\sc i\kern-.025em b}\kern-.08em
    T\kern-.1667em\lower.7ex\hbox{E}\kern-.125emX}}

\usepackage{tikz} \usepackage{lipsum} \makeatletter \def\ieeecopyright{ \footnotesize © 2024 IEEE. Personal use of this material is permitted.\newline DOI: 10.1109/DSD64264.2024.00050} \makeatother \AddToHook{shipout/firstpage}{\begin{tikzpicture}[remember picture,overlay] \node[anchor=south west,xshift=1.0cm,yshift=0.8cm] at (current page.south west){\parbox{\linewidth}{\raggedright\ieeecopyright}}; \end{tikzpicture} }

\begin{document}

\title{AUTOSAR~AP and ROS~2 Collaboration Framework}

\newcommand{\linebreakand}{%
  \end{@IEEEauthorhalign}
  \hfill\mbox{}\par
  \mbox{}\hfill\begin{@IEEEauthorhalign}
}
\makeatother

\author{\IEEEauthorblockN{Ryudai Iwakami} \\\vspace{-2mm}
\IEEEauthorblockA{\textit{Graduate School of}\\\textit{Science and Engineering}\\\textit{Saitama University}} \\\vspace{-10mm}
\and
\IEEEauthorblockN{Bo Peng} \\\vspace{-2mm}
\IEEEauthorblockA{\textit{EMB IV} }\\\vspace{-10mm}
\and
\IEEEauthorblockN{Hiroyuki Hanyu and Tasuku Ishigooka} \\\vspace{-2mm}
\IEEEauthorblockA{\textit{Technology Development Functional Division}\\\textit{Hitachi Astemo, Ltd.}} \\\vspace{-10mm}

\and
\IEEEauthorblockN{Takuya Azumi} \\\vspace{-2mm}
\IEEEauthorblockA{\textit{Graduate School of}\\\textit{Science and Engineering}\\\textit{Saitama University}} \\\vspace{-10mm}
}

\maketitle
\vspace{-7mm}

\begin{abstract}
The field of autonomous vehicle research is advancing rapidly, necessitating platforms that meet real-time performance, safety, and security requirements for practical deployment. 
AUTOSAR Adaptive Platform (AUTOSAR~AP) is widely adopted in development to meet these criteria; however, licensing constraints and tool implementation challenges limit its use in research. 
Conversely, Robot Operating System~2 (ROS~2) is predominantly used in research within the autonomous driving domain, leading to a disparity between research and development platforms that hinders swift commercialization. 
This paper proposes a collaboration framework that enables  AUTOSAR~AP and ROS~2 to communicate with each other using a Data Distribution Service for Real-Time Systems (DDS). 
In contrast, AUTOSAR~AP uses Scalable service-Oriented Middleware over IP (SOME/IP) for communication. 
The proposed framework bridges these protocol differences, ensuring seamless interaction between the two platforms. 
We validate the functionality and performance of our bridge converter through empirical analysis, demonstrating its efficiency in conversion time and ease of integration with ROS~2 tools.
Furthermore, the availability of the proposed collaboration framework is improved by automatically generating a configuration file for the proposed bridge converter.
\end{abstract}


\section{Introduction}

Autonomous driving technology~\cite{Autoware} has rapidly advanced, drawing worldwide attention and spurring significant research and development efforts. 
This surge in activity has brought autonomous vehicles closer to widespread practical use for transporting both people and goods. 
As these vehicles approach commercial readiness, consumer expectations and the landscape of automotive development are evolving. 
Research in this domain is dynamic, propelled by technological advancements and market demand. 
Autonomous driving systems must process data from numerous sensors and cameras in real time, necessitating high technical competence. 
These systems require not only real-time processing capabilities but also flexibility, safety, and security to be viable in practical scenarios.

AUTOSAR Adaptive Platform (AUTOSAR~AP)~\cite{AUTOSARAP,AUTOSAR_AP_OPC_UA} is the chosen platform for addressing the requirements of automotive embedded systems, offering a software architecture that ensures compatibility and extensibility across various vendors and applications. 
Its adoption promotes a standardized approach, revolutionizing automotive development. 
Internet of Vehicles (IoV) strategy enhances this by enabling cloud-based development, testing on actual devices, and iterative feedback-driven enhancements. 
Additionally, AUTOSAR~AP supports Software-Defined Vehicles (SDV) concept, which allows for post-purchase software updates to add new functionalities and value, contrasting with traditional vehicles' static features.
The framework for integrating these ideas and accelerating the development of software-defined vehicles is called SOAFEE~\cite{SOAFEE,SOA_Orchestration}.
Following the SOAFEE trend, AUTOSAR~AP has become an important element in the development of autonomous vehicles.

AUTOSAR~AP is gaining traction as the norm for developing actual autonomous vehicles.
However, its recent establishment means that many current studies opt for alternative platforms, resulting in limited research utilizing AUTOSAR~AP.
Another reason for the lack of research using AUTOSAR~AP is that the use of AUTOSAR~AP requires an expensive license. 
Another, SOME/IP that is a communication protocol used only for in-vehicle networks, is used as a standard.
Given the trend of SOAFEE, it is expected that AUTOSAR~AP and other platforms will be developed in a mixed manner, requiring collaboration and communication with other platforms.
Therefore, support for other communication protocols is desired.
A few existing studies have attempted to communicate between SOME/IP and other communication protocols.

This paper proposes an AUTOSAR~AP and ROS~2 collaboration framework that enables communication between SOME/IP and Data Distribution Service for Real-Time Systems (DDS)~\cite{DDSOMG}.
ROS~2~\cite{ROS2} employs DDS for communication between nodes, the smallest execution units, and uses the same publish/subscribe model as SOME/IP.
ROS~2 is also characterized by its open-source nature, which means that a wealth of documentation, evaluation, and visualization tools are available; this abundance of tools is another reason why ROS~2 is actively used in research. 
The bridge converter proposed in this paper enables the integration between AUTOSAR~AP and ROS~2, which makes it possible to develop AUTOSAR~AP using the comprehensive ROS~2 tools.
ROS~2 platform can also be used more effectively as a standard platform for actual vehicle development. 
This synergy aids the rapid transition of ROS~2 research findings into the commercial development of autonomous vehicles.
Furthermore, it contributes to the realization of SOAFEE, which is expected to be developed with a mixture of AUTOSAR~AP and ROS 2, and facilitates the development of autonomous vehicles.
 
The primary contributions of this paper are summarized as follows:
 \begin{itemize}
\item \textbf{Facilitating AUTOSAR~AP and ROS~2 collaboration:} The proposed bridge converter performs DDS and SOME/IP conversion between AUTOSAR~AP and ROS~2, allowing communication between the two platforms.
\item \textbf{Enabling SOME/IP data visualization:} The proposed bridge converter enables the use of ROS~2's rich evaluation and visualization tools for AUTOSAR~AP applications.
\item \textbf{Supporting diverse custom messages: }By entering the necessary ID and name details for SOME/IP communication, our system auto-generates the configuration file, simplifying the use of our bridge converter.
 \end{itemize}

The remainder of this paper is organized as follows. 
Section~\ref{chap:system_model} describes the characteristics of AUTOSAR~AP and ROS~2, which are the prerequisites for this study. 
Section~\ref{chap:design_and_implementation} describes the proposed communication method. 
Section~\ref{chap:evaluation} describes the evaluation of the proposed communication method, and Section~\ref{chap:related_work} describes related work. Finally, a brief conclusion is given. 

\vspace{-3mm}
\section{System Model}
\label{chap:system_model}

 \begin{figure}[t]
\centering
     \includegraphics[width=7truecm]{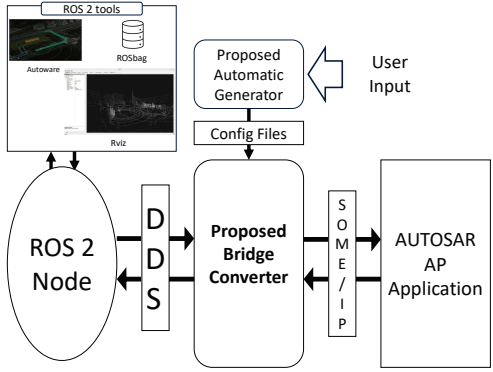}
    \caption{The rough architecture proposed in this study.}
    \label{fig:abstract_architecture}
    \vspace{-3mm}
\end{figure}

This section presents an overview of the proposed method and explains AUTOSAR~AP and ROS~2, which are considered in the proposed method. 
An overview of the proposed method is summarized in Fig.~\ref{fig:abstract_architecture}. 
The proposed method implements a bridge that converts the communication method between the ROS~2 node and the AUTOSAR~AP application for communication between each component. 
This bridge resolves the differences between DDS and SOME/IP and converts the communication between the two.The following is a prerequisite knowledge of the proposed method.

Firstly, the structure and features of another platform, AUTOSAR~AP, are described.
Secondly, the structure of ROS~2 and an overview of the communication schemes implemented are described.
Thirdly, SOME/IP, the communication protocol mainly used in AUTOSAR~AP, is explained. 
Fourthly, DDS used in ROS~2 is explained. 
Finally, vsomeip and CommonAPI used in this implementation are described.

\subsection{AUTomotive Open System ARchitecture  Adaptive Platform (AUTOSAR~AP)}
\label{sec:AUTOSARAP}

AUTOSAR~AP~\cite{comparison} is a software architecture for next-generation automotive embedded systems. 
Compared to ROS~2 with DDS, AUTOSAR~AP uses SOME/IP as its main communication protocol. 
Besides, while ROS~2 is open-source, AUTOSAR~AP requires a license for use and is not freely available. 
Therefore, AUTOSAR~AP is not as richly documented as ROS~2 and has few existing results.
Another feature is the difficulty of handling due to excessive standardization.

Similar to the design of ROS~2~\cite{comparison}, the architecture of AUTOSAR~AP can be classified into three layers: the application layer, the middleware layer, and the OS layer. 
The application layer includes the application and Non-Platform (Non-PF) Service, where services specific to a particular application or system are defined by the user. 
The middleware layer provides the basic functionalities and services of AUTOSAR~AP and is called AUTOSAR~Runtime for Adaptive Application (ARA). POSIX-based OSs can be used as the OS layer. 
For example, multiple operating systems such as Linux, QNX, and INTEGRITY can be used.

The communication flow of an AUTOSAR~AP is shown in Fig.~\ref{fig:AUTOSAR_AP_simple_architecture}. 
AUTOSAR~AP uses a Service-Oriented Architecture (SOA), and the nodes in ROS~2 are called applications. 
Applications are provided as services, and communication with other applications and components is done as service requests. SOME/IP is a protocol that supports service-oriented communication, enabling dynamic service discovery and service requests. 
Next, another platform, ROS~2, is described in comparison with AUTOSAR~AP.

 \begin{figure}[t]
\centering
     \includegraphics[width=4.5truecm]{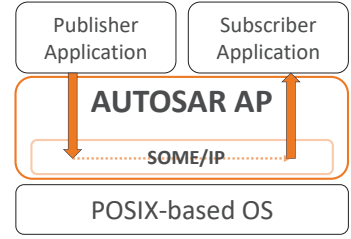}
     \caption{Communication between AUTOSAR~AP applications.}
    \label{fig:AUTOSAR_AP_simple_architecture}
\end{figure}

 \begin{figure}[t]
\centering
\vspace{-5mm}

     \includegraphics[width=4.5truecm]{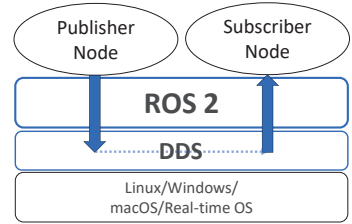}
    \caption{Communication between ROS~2 nodes.}
    \label{fig:ROS_2_simple_architecture}
       \vspace{-3mm}
\end{figure}

\subsection{Robot Operating System 2 (ROS~2)}
\label{sec:ROS2}
ROS~2~\cite{comparison}, an open-source middleware, finds use in autonomous driving software. 
This platform represents an improved version of ROS~and frequently appears in industrial applications due to its ability to meet the strict real-time constraints inherent to such settings. 
In contrast to ROS, ROS~2's most notable feature is the incorporation of DDS for real-time processing. 
With ROS~2's open-source nature and broad application across industries, researchers have access to a plethora of tools for efficient study.

The architecture of applications based on ROS~2 comprises three layers: the application layer, the middleware layer, and the OS layer. 
The application layer has user code and ROS~2 nodes. Internal processing of each ROS~2 node occurs in the middleware layer, with DDS handling communication between nodes. 
This architecture supports multiple operating systems, including Linux, Windows, macOS, and real-time OSs.

The communication flow of ROS~2 is shown in Fig.~\ref{fig:ROS_2_simple_architecture}. 
ROS~2 creates nodes as units of communication, and each node exchanges information with each other using a data bus (hereafter referred to as ``topic"). 
DDS is discussed in detail in Section~\ref{sec:DDS} below.

 \begin{figure}[t]
\centering
\vspace{-2mm}

     \includegraphics[width=7truecm]{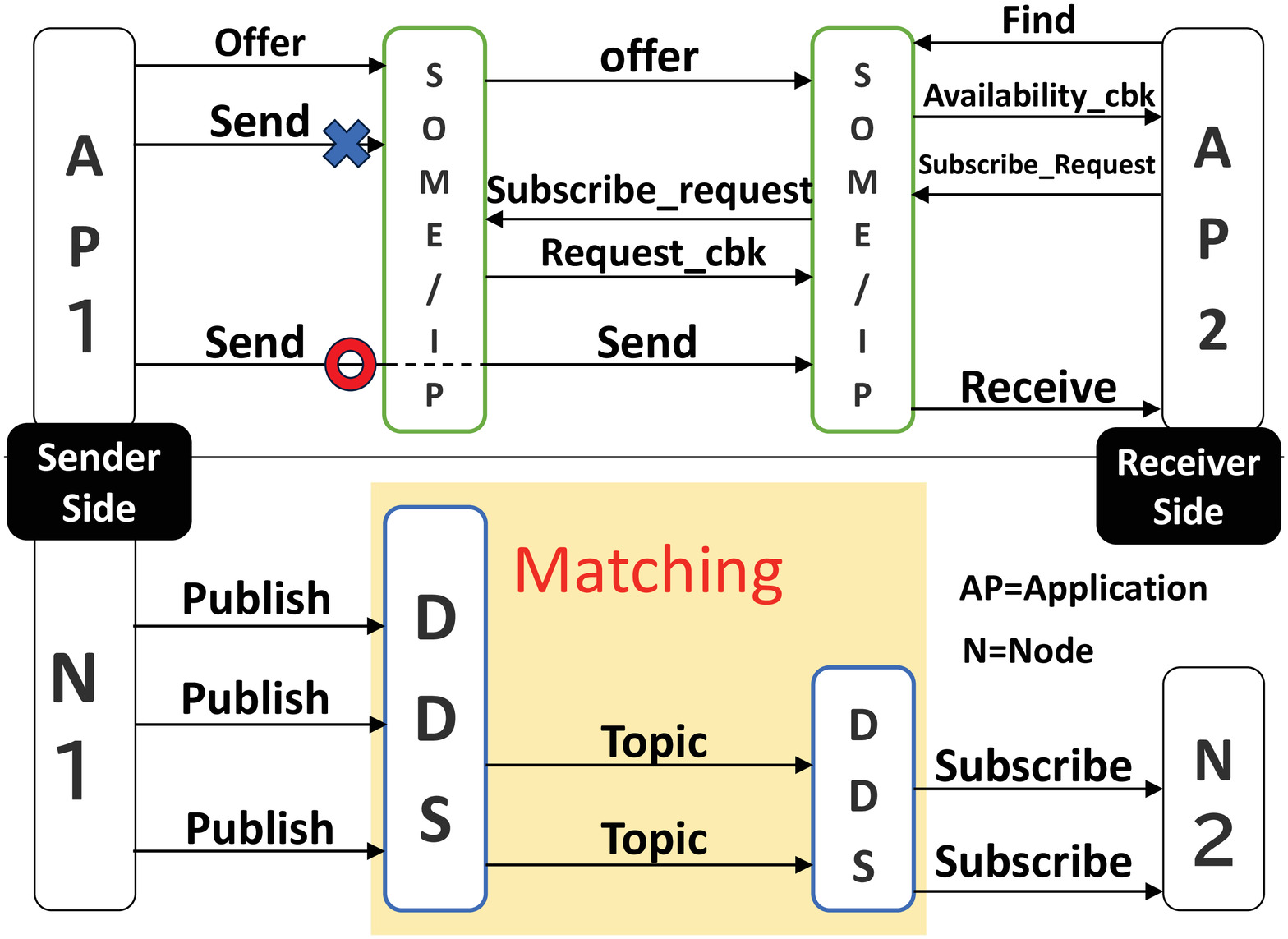}
    \caption{Comparison of DDS and SOME/IP send/receive sequences.}
     \label{fig:DDS_SOMEIP}
     \vspace{-5mm}
\end{figure}

\subsection{Scalable service-Oriented MiddlewarE over IP (SOME/IP)}
\label{sec:SOME/IP}

SOME/IP~\cite{gateway} is a communication protocol specialized for automotive networks among real-time distributed systems and enables service-oriented communication over Ethernet. 
SOME/IP incorporates SOME/IP Service Discovery (SOME/IP-SD), which can dynamically discover and register services on the network. 
SOME/IP-SD uses service-specific identifiers to discover and register services. ServiceID and InstanceID are used as identifiers.
ServiceID is an identifier to identify a specific service uniquely.
InstanceID is an identifier to identify different instances of the same service when there are multiple instances of the same service.

The flow of SOME/IP-SD is shown in the upper part of Fig.~\ref{fig:DDS_SOMEIP}.
First, the sending application sends an Offer message, and the receiving application sends a Find message.
Next, when the offered function matches the function being searched for, the receiver sends a subscribe message and data is exchanged.
SOME/IP messages are divided into a header part and a payload part. The payload part contains the data to be sent and received.
The header part of a SOME/IP message contains SOME/IP's unique identifiers, such as ServiceID and InstanceID, and the information in the header part is used to match the functionality of the message.
``Subscribe\_request" in Fig.~\ref{fig:DDS_SOMEIP} because the nuance of Publish/Subscribe in SOME/IP is slightly different from that in ROS~2. In addition, ``cbk" is an abbreviation for ``callback."
The difference from the DDS described in the next section is that the application side recognizes the matching part by checking the existence of the receiver. The following section describes DDS used in ROS~2.

 \begin{figure}[t]
\centering
     \includegraphics[width=6truecm]{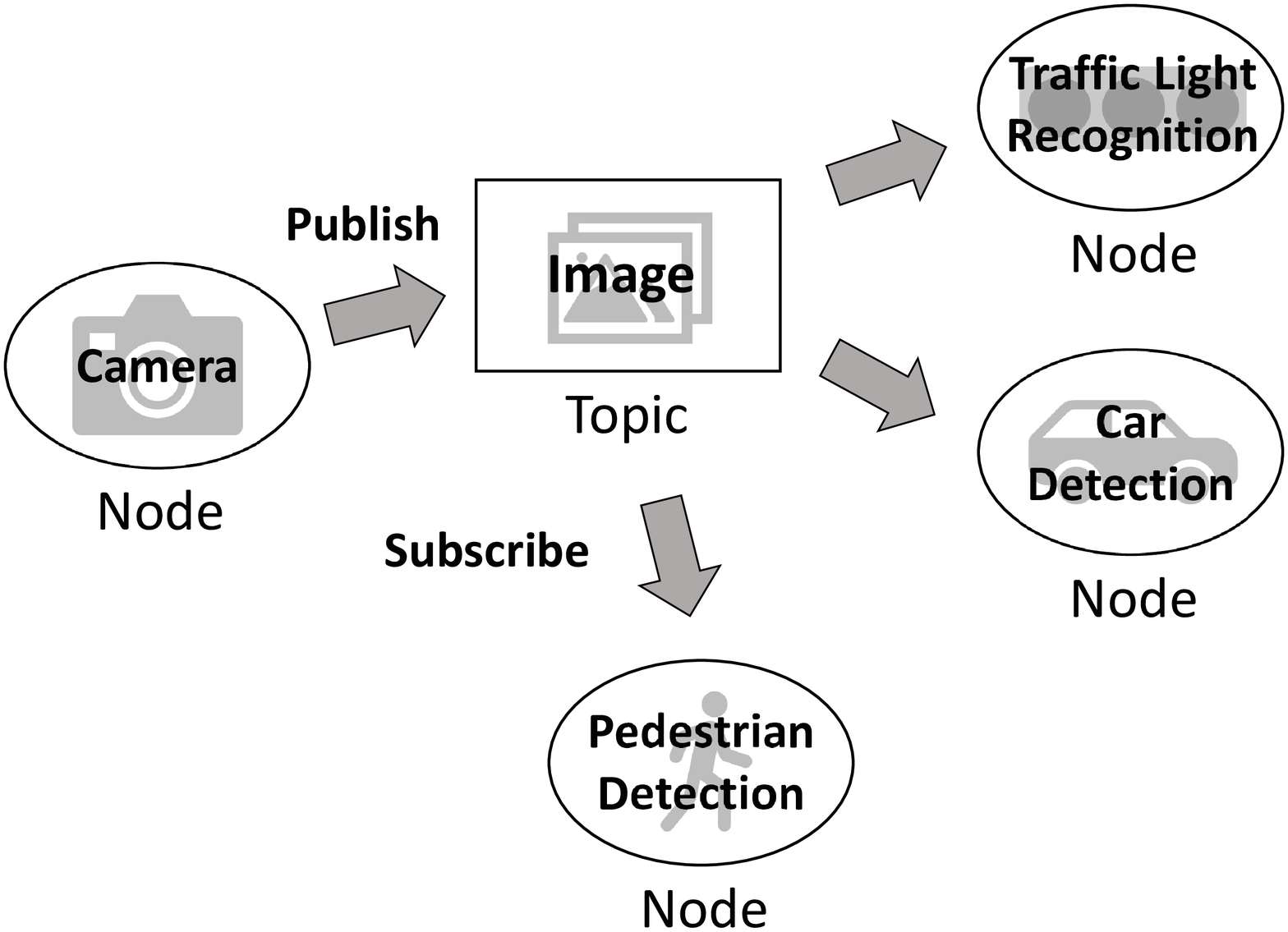}
    \caption{Publish/Subscribe model.}
     \label{fig:Publish/Subscribe}
        \vspace{-3mm}
\end{figure}

\subsection{Data Distribution Service for Real-Time Systems (DDS)}
\label{sec:DDS}

DDS~\cite{gateway,ROS2} is a middleware protocol for real-time distributed systems, especially fitting for large and demanding systems.
The main feature of DDS is that it employs a data-centric communication model, which is called the Publish/Subscribe model.
The flow of the Publish/Subscribe model is shown in Fig.~\ref{fig:Publish/Subscribe}. In the publish/subscribe model, data is exchanged via a data bus called a topic.
Sending a topic is called publishing, and receiving a topic is called subscribing.
When a topic is published, the publisher node can communicate without specifying a subscriber node, and information is exchanged dynamically via topics.
In other words, the middleware manages the matching so that a sending node can publish information to a topic regardless of whether or not a receiving node is present.
The lower part of Fig.~\ref{fig:DDS_SOMEIP} shows the flow of sending and receiving DDS.
The topic type is statically described in the node's internal source code. 
When a publisher node is invoked, a topic of the type based on the node's description is published. 
The published data is managed by the DDS middleware, and when the appropriate subscriber node is invoked, the topic is subscribed, and data reception is complete.
The difference with SOME/IP is that a node is not aware of its communication partners.

\section{Design and Implementation}
\label{chap:design_and_implementation}

This section introduces the DDS-SOME/IP conversion tool implemented in this study. 
First, the structure of the tool, as shown in Fig.~\ref{fig:abstract_architecture}, is described. 
Then, specific implementation techniques are described.

\subsection{Structure of DDS-SOME/IP Bridge Converter}
\label{sec:Structure of DDS-SOME/IP Bridge Converter}

This study implemented a bridge facilitating communication between AUTOSAR~AP applications and ROS~2 nodes by performing DDS and SOME/IP conversion, as illustrated in Fig.~\ref{fig:Bridge_on_ROS2}. 
The two main functionalities of the bridge are:
\begin{itemize}

\item  \textbf{Support for SOME/IP-SD: }Since SOME/IP has a service discovery functionality that DDS does not have, the bridge performs SOME/IP-SD process to establish communication.
\item  \textbf{Type conversion: }Since AUTOSAR~AP and ROS~2 use different types for data exchange, the bridge performs conversion to a mutually recognized type.
 \end{itemize}

\begin{figure}[t]
\centering
     \includegraphics[width=6.5truecm]{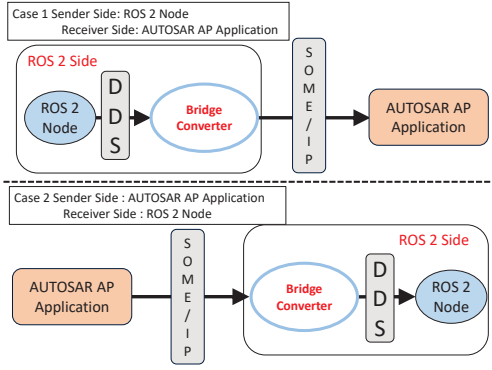}
    \caption{Two cases of the bridge as ROS~2 node.}
     \label{fig:Bridge_on_ROS2}
         \vspace{-3mm}
\end{figure}

 The data exchange between a ROS~2 node and an AUTOSAR~AP application can be divided into two cases according to the direction of data flow, as shown in Fig.~\ref{fig:Bridge_on_ROS2}. 
 The upper side represents the case where ROS~2 node is the sender and the AUTOSAR~AP application is the receiver. 
 The lower side represents the case where the AUTOSAR~AP application is the sender and ROS~2 node is the receiver. 
 The bridge converter implemented in this study was implemented as a ``node" on the ROS~2 side in both cases.
 By implementing a bridge converter as a ``node" on the ROS~2 side, ROS~2 tools such as Rviz~\cite{Rviz} and ROSbag~\cite{rosbag} can be easily used. 
 This method is easier to achieve the goal of this study, which is to link AUTOSAR~AP and ROS~2, than other methods.

In this study, the bridge was implemented as a ``node" on the ROS~2 side, yet alternative methods exist for linking AUTOSAR~AP and ROS~2. 
One of the methods is to implement a component with bridge functionality as an “application” on the AUTOSAR~AP side.
However, as mentioned in Section 2.B, AUTOSAR~AP requires a license, so there are fewer existing achievements compared to ROS~2.
Implementing the bridge converter on the AUTOSAR~AP side would be less scalable than implementing the bridge converter on the ROS~2 side.
On the other hand, ROS~2 is open source and has many tools and libraries.
Therefore, implementing the bridge converter as a ``node" rather than as an ``application" is considered to increase the scalability of the bridge converter.
Another possible approach is to make AUTOSAR~AP compatible with DDS.
The AUTOSAR~AP specification document describes support for DDS.
However, most of the descriptions of DDS are drafts and only partially supported in practice~\cite{AUTOSAR_text}.
With only a partial specification, to make AUTOSAR~AP compatible with DDS will be a future technical debt.
For this reason, this method was not selected for this study. 
With the above background, this study implemented a component with bridge functionality as a ``node" on the ROS~2 side.

\begin{table}[t]
    \centering

    \caption{Requirements for generation FIDL files}
    \begin{tabular}{|c|c|}
    \hline
    Requirement & Description \\ \hline 
    Interface name & Name for the interface \\ \hline
    Method name & Name for the method \\ \hline
    Major version & Main version of the service's API \\ \hline
    Minor version &  Subversion of the service's API \\ \hline
    Service ID &  Unique identifier for each service \\ \hline
    Instance ID & Unique identifier for each instance \\ \hline
    Event ID & unique identifier for each event \\ \hline
    Eventgroup ID & unique identifier for each eventgroup \\ \hline
    Service name & Name for the service \\ \hline
    Package name & Name for the package \\ \hline
    \end{tabular}
    \label{tab:automatic_generator}
\vspace{-4mm}
\end{table}

 \begin{figure}[t]
\centering
     \includegraphics[width=7truecm]{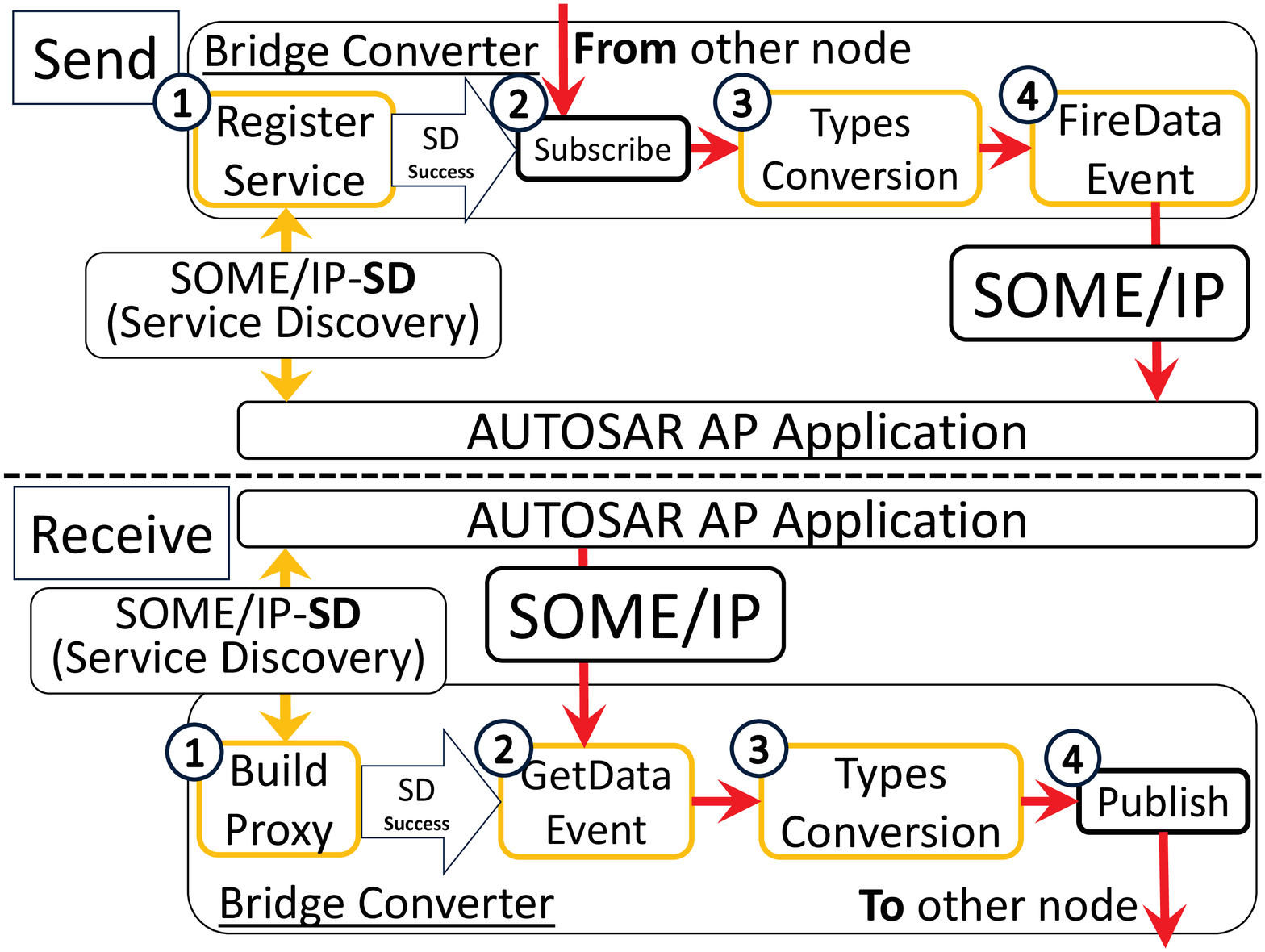}
    \caption{Proposed bridge converter implementation.}
     \label{fig:bridge_converter_imple}
\vspace{-4mm}
\end{figure}

\subsection{Implementation Method of the Proposed Bridge Converter}
\label{sec:Implementation method of the proposed bridge converter}

To implement the proposed bridge converter, CommonAPI~\cite{GNSSbridge} and vsomeip~\cite{VSOMEIP,GNSSbridge} were used. 
CommonAPI supports development as an AUTOSAR standard and provides a high-level API to facilitate application development for automotive systems. 
Its compliance with the AUTOSAR standard and its open-source nature were the reasons for its adoption in the implementation of this study. 
vsomeip is also one of the open-source implementations of SOME/IP protocol used in the automotive industry to facilitate communication between different devices in a vehicle. 

Since the implementation adheres to the industry-standard SOME/IP protocol, this approach offers superior compatibility and standardization. 
The open-source nature of the protocol influenced its selection for use in this study's implementation.
CommonAPI uses FIDL (Franca Interface Definition Language) files and FDEPL (Franca Deployment) files to define the service interface, which describe the methods provided by the service, events to be sent, and attributes to be exposed. They also describe information about type conversion. 
The schematic of the bridge converter implementation is shown in Fig.~\ref{fig:bridge_converter_imple}. 
The yellow components represent vsomeip functions, and the red arrows represent data flow. As shown in Fig~\ref{fig:Bridge_on_ROS2}, the proposed bridge converter can be divided into two patterns. First, when ROS~2 is the sender, the RegisterService function performs SOME/IP-SD based on the information in the FIDL file. When a sender is found, it subscribes to the data from the predecessor node and performs type conversion. Finally, the data is sent as a SOME/IP message using the FireDataEvent function. Next, if ROS~2 is the receiver, the BuildProxy function performs SOME/IP-SD based on the information in the FIDL file. When a receiver is found, it receives the data using the GetDataEvent function and performs type conversion. Finally, the data is published to the successor node.

\subsection{FIDL Automatic Generator}

SOME/IP requires the matching of information, such as ServiceID, between the server and client sides. The specifications are all described in ARXML files, and the settings of identification information such as ServiceID are also described in ARXML files. In contrast, the proposed bridge converter using CommonAPI uses FIDL and JSON files to set the identification information. Since the identification information for both files needs to be aligned, the FIDL and JSON files need to be re-generated each time the AUTOSAR~AP application changes. Therefore, the collaboration framework proposed in this study implements the FIDL Automatic Generator, which can automatically generate FIDL files.

The FIDL Automatic Generator is a GUI-based program using Python.
The FIDL file and FDEPL file that can be used with the bridge converter can be generated automatically by entering the items shown in Table~\ref{tab:automatic_generator}.
The FIDL Automatic Generator can automatically generate tens of lines of FIDL and FDEPL files with only a few lines of input, facilitating the use of the proposed bridge converter.
In addition, by loading msg files that define custom messages, a FIDL file describing the conversion specifications can be automatically generated.
For example, when dealing with sensor\_msgs::msg::PointCloud2 type messages, the file generated for the conversion is 78 lines long. Using the proposed tool, a configuration file can be generated automatically with only a few entries.

\section{Evaluation}
\label{chap:evaluation}

This section summarizes the delay evaluation of the DDS-SOME/IP conversion bridge converter implemented in this study by message size. 
In addition, the evaluation using Rviz as a ROS~2 evaluation tool and the extension of the development method using the bridge converter is presented. 
First, the environment used for the evaluation is described.

\subsection{Evaluation Environment}
\label{sec:Evaluation Environment}

 \begin{figure}[t]
\centering

     \includegraphics[width=6truecm]{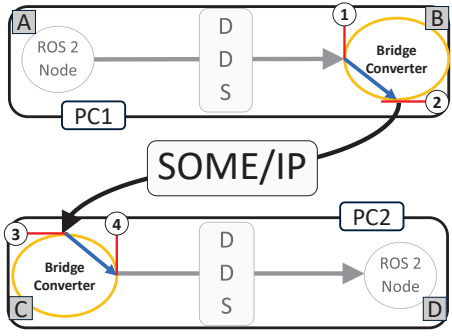}
    \caption{Evaluation program structure.}
     \label{fig:Evaluation_program_structure_number}
\end{figure}

\begin{figure}[t]
\centering
\vspace{-3mm}
     \includegraphics[width=6truecm]{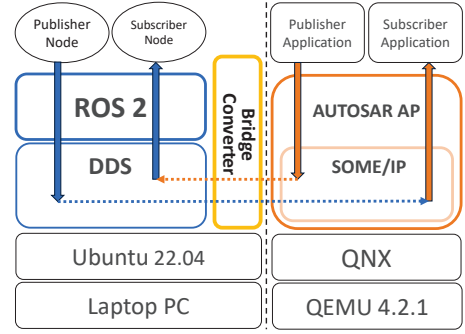}
    \caption{Architecture of communication between AUTOSAR~AP and ROS~2 in this paper.}
    \label{fig:architecture of HW levels}
       \vspace{-4mm}
\end{figure}

\begin{table}[t]
    \centering
    \vspace{-3mm}
    \caption{Environment on PC1}
    \setlength{\tabcolsep}{2pt} 
    \begin{tabular}{|c|c|c|}
    \hline
    Hardware & Processor& Intel\textregistered Core i9-12900HX CPU@2.3 GHz\\ \hline
     & Memory& 32~GB \\ \hline
    OS & Ubuntu & 22.04.3 \\ \hline
    Middleware & ROS~2 & Humble Hawksbill \\ \hline
         \scriptsize Communication  Protocol &

         DDS      &   Fast DDS \\ \hline
          & SOME/IP & vsomeip 3.4.9.1~\cite{SOMEIP} 
    \\ \hline
  
\end{tabular}

    \label{tab:Env_ROS2}
      \vspace{-4mm}
\end{table}

\begin{table}[t]
    \centering

    \caption{Environment on PC2}
    \setlength{\tabcolsep}{2pt} 
    \begin{tabular}{|c|c|c|}
    \hline
    Hardware & Processor& Intel\textregistered Core i7-10850H CPU@2.3 GHz\\ \hline
     & Memory& 16~GB \\ \hline
    OS & Ubuntu & 22.04.3 \\ \hline
    Middleware & ROS~2 & Humble Hawksbill \\ \hline
         \scriptsize Communication  Protocol &

         DDS      &   Fast DDS \\ \hline
          & SOME/IP & vsomeip 3.4.9.1~\cite{SOMEIP} 
    \\ \hline
    \end{tabular}
        \vspace{-5mm}
    \label{tab:Env_ROS2_2}
\end{table}

\begin{table}[t]
    \centering
    \caption{Environment of AUTOSAR~AP on QEMU}
    \setlength{\tabcolsep}{1pt} 
    \begin{tabular}{|c|c|c|}
    \hline
    Hardware & Processor& Intel\textregistered Core i9-12900HX CPU@2.3 GHz\\ \hline
     & Memory&  32~GB \\ \hline
    VM & QEMU & 4.2.1 \\ \hline
     & Memory & 4~GB \\ \hline
    OS & QNX & \\ \hline
    Middleware & AUTOSAR~AP & R20-11 \\ \hline
    \begin{tabular}{c}
         Communication \\ Protocol
    \end{tabular} & SOME/IP & \\ \hline
    \end{tabular}
        \vspace{-3mm}
    \label{tab:Env_AUTOSARAP}
\end{table}

First of all, the environment used to evaluate the latency of the DDS-SOME/IP bridge converter by message size is shown in Fig.~\ref{fig:architecture of HW levels}.
In this evaluation, PC1 shown in Table~\ref{tab:Env_ROS2} and PC2 shown in Table~\ref{tab:Env_ROS2_2} were used to execute the evaluation program shown in Fig.~\ref{fig:Evaluation_program_structure_number} to evaluate latency. 
Since the default in ROS~2 humble is FastDDS, FastDDS is used in the evaluation in this study. For CycloneDDS, we confirmed that the proposed bridge converter works.
In addition, in the latency evaluation of the communication with AUTOSAR~AP applications, as shown in Fig.~\ref{fig:architecture of HW levels}, PC1 and Table~\ref{tab:Env_AUTOSARAP} were used via the proposed bridge converter. The latency was evaluated by communicating with applications in the AUTOSAR~AP environment as shown in Fig.~\ref{fig:Evaluation_program_structure_number}, through the proposed bridge converter.
QNX and AUTOSAR~AP were built on QEMU, and a sending application and a receiving application were created as AUTOSAR~AP applications.
In the evaluation of this paper, ROS~2 is used on an actual device, but AUTOSAR~AP is used on QEMU. This is because, although AUTOSAR~AP is guaranteed to work on POSIX OS, it is not guaranteed to work on general development environments (e.g., Linux, Windows). Therefore, in the function development stage, it is common to use an environment such as QEMU.

\subsection{Latency Evaluation of DDS-SOME/IP Conversion Tool}
\label{sec:Latency Evaluation of DDS-SOME/IP Conversion Tool}

This section describes the latency evaluation results of the proposed bridge converter. 
The structure of the evaluation program and the six checkpoints are shown in Fig.~\ref{fig:Evaluation_program_structure_number}. 
The details of the checkpoints are summarized in Table~\ref{tab:checkpoints}.
Section~\ref{421} provides an overview of the simple program used for the evaluation, in which both SOME/IP and DDS are used. 
The delay evaluation of the proposed bridge converter is based on two possible use cases: in Section~\ref{subsec:Latency to Convert and Send PointCloud2 Type Data from DDS to SOME/IP}, the measurements are based on the case where the sender is a ROS~2 node and the receiver is an AUTOSAR~AP application, as shown in the upper part of Fig.~\ref{fig:Bridge_on_ROS2}. 
Section~\ref{sebsec:Latency to  Convert and Receive PointCloud2 Type Data from SOME/IP to DDS} is based on the case where the sender is an AUTOSAR~AP application, and the receiver is a ROS~2 node, as shown in the lower part of Fig.~\ref{fig:Bridge_on_ROS2}.

\subsubsection{Program Overview for Evaluation}
\label{421}
The point cloud data of sensor\_msgs::msg::PointCloud2 type was used for the evaluation, and measurements were performed on seven different data sizes with different numbers of point clouds. 
ROS~2 node A generates sensor\_msgs::msg::PointCloud2 type data and sends the data to bridge converter node B via DDS. 
The bridge converter node at C receives the data, converts the data to sensor\_msgs::msg::PointCloud2 type based on the description in the FIDL file, and sends the data to the ROS~2 node at D through DDS. Finally, the ROS~2 node in D receives and displays the data. 
The above program was executed in the environment shown in Tables~\ref{tab:Env_ROS2} and~\ref{tab:Env_ROS2_2}.

\begin{table}[t]
    \centering
    \caption{Check point for evaluation}
    \setlength{\tabcolsep}{2pt} 
    \begin{tabular}{|c|l|}
    \hline
    Check point number & Description \\ \hline
     $\textcircled{\scriptsize 1}$ &  Just before DDS to SOME/IP conversion\\
     \hline
     $\textcircled{\scriptsize 2}$ &    \begin{tabular}{l}
          Immediately after DDS to SOME/IP conversion \\ (Immediately before SOME/IP sending)
    \end{tabular}\\ \hline
   $\textcircled{\scriptsize 3}$
    
         &  \begin{tabular}{l}
          Immediately after receiving SOME/IP \\ (Immediately before SOME/IP to DDS conversion)
    \end{tabular}\\ \hline
   $\textcircled{\scriptsize 4}$
  
         &  Immediately after  SOME/IP to DDS conversion\\
 \hline 
    \end{tabular}
    \vspace{-3mm}
    \label{tab:checkpoints}
\end{table}

\subsubsection{Latency to Convert and Send PointCloud2 Type Data from DDS to
SOME/IP}
\label{subsec:Latency to Convert and Send PointCloud2 Type Data from DDS to SOME/IP}

 \begin{figure}[t]
\centering
     \includegraphics[width=7truecm]{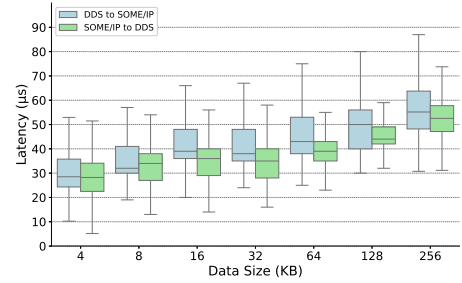}
    \caption{Latency to conversion.}
     \label{fig:DtoS}
         \vspace{-5mm}
\end{figure}

 \begin{figure}[t]
\centering
     \includegraphics[width=7truecm]{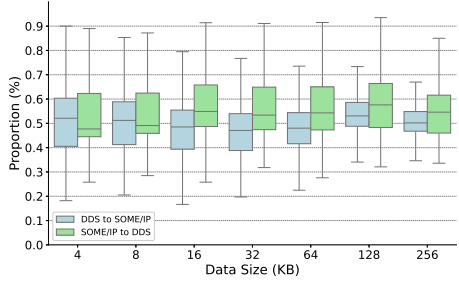}
    \caption{Proportion of conversion time to communication time.}
     \label{fig:PropDtoS}
         \vspace{-5mm}
\end{figure}

The measurements in this section are based on the case shown in the upper part of Fig.~\ref{fig:Bridge_on_ROS2}, where the sender is a ROS~2 node, and the receiver is an AUTOSAR~AP application.
This case requires conversion from DDS to SOME/IP.
Therefore, the evaluation in this section uses the time from $\textcircled{\scriptsize 1}$ to $\textcircled{\scriptsize 2}$ in Fig.~\ref{fig:Evaluation_program_structure_number}. 
The latency was measured for each number of point clouds and is summarized in the box-and-whisker diagram on the left side of Fig.~\ref{fig:DtoS}.
The results show that the latency increases as the data size increases. However, the latency is kept below 100$~\mu$s for all data sizes, indicating that the latency caused by the conversion from DDS to SOME/IP is small.

Using this measurement result and the time taken from $\textcircled{\scriptsize 2}$ to $\textcircled{\scriptsize 3}$ in Fig.~\ref{fig:Evaluation_program_structure_number}, we can calculate $\textcircled{\scriptsize 1}$ to $\textcircled{\scriptsize 3}$ was used to calculate the ratio of the execution time of the $\textcircled{\scriptsize 1}$ to $\textcircled{\scriptsize 2}$ part. 
The result of calculating the percentage of execution time is shown in the box-and-whisker diagram on the left side of Fig.~\ref{fig:PropDtoS}.
Most of the values are within 1\% or less, indicating that the time spent for conversion is a small latency compared to the communication latency. Furthermore, the ratio does not vary significantly with data size, suggesting that it increases at the same rate as the communication latency.

\subsubsection{Latency to Convert and Receive PointCloud2 Type Data from
SOME/IP to DDS}
\label{sebsec:Latency to  Convert and Receive PointCloud2 Type Data from
SOME/IP to DDS}
The measurements in this section are based on the case shown in the lower part of Fig.~\ref{fig:Bridge_on_ROS2}, where the sender is an AUTOSAR~AP application, and the receiver is a ROS~2 node. This case requires conversion from SOME/IP to DDS. Therefore, the evaluation in this section uses the time from $\textcircled{\scriptsize 3}$ to $\textcircled{\scriptsize 4}$ in Fig.~\ref{fig:Evaluation_program_structure_number}. The latency was measured for each number of point clouds and is summarized in the box-and-whisker diagram on the right side of Fig.~\ref{fig:DtoS}. The results show that the latency increases as the data size increases, similar to the latency for the conversion from DDS to SOME/IP. The latency of the conversion from SOME/IP to DDS is small.

Using these measurements and the time taken from $\textcircled{\scriptsize 2}$ to $\textcircled{\scriptsize 3}$ in Fig.~\ref{fig:Evaluation_program_structure_number}, the proportions of the conversion time to the communication time were calculated.
The box-and-whisker diagram on the left side of Fig.~\ref{fig:PropDtoS} shows the result of calculating the ratio of the execution time of parts $\textcircled{\scriptsize 2}$ to $\textcircled{\scriptsize 4}$ to the time taken from $\textcircled{\scriptsize 3}$ to $\textcircled{\scriptsize 4}$.
A similar trend to Section~\ref{subsec:Latency to Convert and Send PointCloud2 Type Data from DDS to SOME/IP} is observed, and most of the values are within 1\% or less, indicating that the time spent for conversion is a small latency compared to the latency for communication. Furthermore, this ratio does not vary significantly with data size, suggesting that it increases at the same rate as the communication latency.

    \vspace{-3mm}
\subsection{Latency Evaluation in Communication with AUTOSAR~AP Application}

For the AUTOSAR~AP application, the proposed bridge converter was used to convert DDS to SOME/IP, and 10 types of String type data were sent with different sizes, and the latency was measured. The results are summarized in Fig.~\ref{fig:toAUTOSAR_string}. It can be seen that the measured latency is higher than expected despite the small data size, and that the latency does not increase as the data size increases. Furthermore, considering the results of Section~\ref{subsec:Latency to Convert and Send PointCloud2 Type Data from DDS to SOME/IP}, it is considered that the communication bottleneck is not the bridge converter but other parts such as QEMU.

 \begin{figure}[t]
\centering
     \includegraphics[width=7truecm]{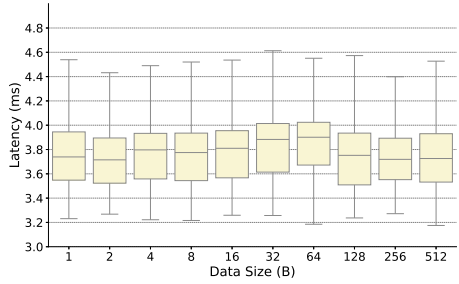}
    \caption{Latency of sending String type data to AUTOSAR~AP applications.}
     \label{fig:toAUTOSAR_string}
         \vspace{-5mm}
\end{figure}

\subsection{Improved Usability of AUTOSAR~AP with ROS 2 Tools}
\label{sec:Improved usability of AUTOSAR AP with ROS 2 tools}

\begin{table}[t]
    \centering

      \label{tab:input}
    \caption{Number of auto output lines and manual input items}
    \small
    \label{tab:input}
    \begin{tabular}{|l|c|cc|c|} \hline
    Message type & Manual inputs &  a & b & Total \\ \hline
    Point.msg & 14 & 51 & 20 & 71\\  \hline
    NavSatFix.msg & 14 & 51 & 24 & 75 \\  \hline
    PointCloud2.msg & 14 & 51 & 27 & 78 \\  \hline
    TFMessage.msg & 14 & 51 & 34 & 85 \\  \hline
    Waypoint.msg & 14 & 51 & 89 & 140 \\ \hline
    \end{tabular}
    \vspace{2mm}
   \\ \textbf{a:} Number of generated lines does not vary by message type \\
   \textbf{b:} Number of generated lines that vary by message type
      \vspace{-5mm}
\end{table}

ROS~2 has a wealth of development support tools, such as Rviz~\cite{Rviz}. 
AUTOSAR~AP does not have such a convenient tool, and the debugger must use Wireshark and other tools to investigate the cause steadily. As a result, the efficiency of the research will be reduced compared to ROS~2.
The proposed bridge converter makes it possible to use the following ROS~2 development support tools even in an environment using SOME/IP communication, such as AUTOSAR~AP.

\subsubsection{Visualization of Communication Using Rviz}
\label{subsec:Visualization of communication using Rviz}
As shown in the upper left corner of Fig.~\ref{fig:abstract_architecture}, the proposed method enables Rviz to be used with AUTOSAR~AP applications as well, which contributes to more efficient research and development by visualizing topics and facilitating understanding of the data being exchanged. 
In this study, we have made it possible to use Rviz to visualize data sent from AUTOSAR~AP applications via the bridge converter. 
Furthermore, in communication with Autoware (Autonomous driving software), data can be visualized through Rviz.
Autoware~\cite{Autoware} is ROS~2-based autonomous driving software widely used in the autonomous driving field. 
The proposed framework enables communication with ROS~2-based software such as Autoware.

\begin{table*}[t]
\centering
    \caption{Related Work and relevant fields}
    \label{tab:related_work}
     \vspace{-2mm}
    \footnotesize
    \begin{tabular}{|l|ccccc|}
    \hline
        &  ROS~2& AUTOSAR~AP& DDS& SOME/IP& SOA  \\ \hline 
             SOA for  autonomous vehicles~\cite{SOC_in_FAN,SOA_for_HAV} 
       & & & &$\checkmark$& $\checkmark$\\ \hline 
             SOA using  dynamic orchestration~\cite{SOA_Orchestration} 
        &   $\checkmark$& &&& $\checkmark$ \\ \hline
            Beginning of AUTOSAR~AP~\cite{AUTOSAR_AV}
        &  & $\checkmark$&  & & \\ \hline 
             Comparison of AUTOSAR~AP and ROS~2~\cite{comparison}
        &   $\checkmark$& $\checkmark$& & & $\checkmark$ \\ \hline
              Use OPC UA on AUTOSAR~AP~\cite{AUTOSAR_AP_OPC_UA}
        &  & $\checkmark$&  & & \\ \hline 
             Gateway for  DDS and OPC~UA~\cite{DDS_OPC_UA}
        &  & & $\checkmark$& & \\ \hline
            Gateway for  SOME/IP and OPC~UA~\cite{SOMEIP_OPC_UA}
        &  & & & $\checkmark$&\\ \hline
            Gateway for DDS, SOME/IP and e-CAL~\cite{gateway}
      &  & $\checkmark$& $\checkmark$& $\checkmark$ \\ \hline
            Bridge between ROS~2 and Adaptive-AUTOSAR~\cite{AUTOSAR_AP_ROS_2}
       & & $\checkmark$& $\triangle$& $\checkmark$& $\triangle$ \\ \hline
        This paper& $\checkmark$ & $\checkmark$ & $\checkmark$ & $\checkmark$ & $\checkmark$ \\ \hline
    \end{tabular}
            \vspace{-4mm}
\end{table*}

\subsubsection{Easier Debugging with ROSbag}
\label{subsec:Easier debugging with Rosbag}

As well as Rviz, the proposed method makes ROSbags available to AUTOSAR~AP applications, allowing topics to be saved and replayed, making it easier to reproduce the same situation. As a result, debugging is much more efficient, contributing to more efficient research and development. As shown in Fig.~\ref{fig:abstract_architecture}, this paper uses ROSbag to store data to be sent to an AUTOSAR~AP application, and the same data can be sent repeatedly without effort via a bridge converter. As a result, it contributes to the improvement of debugging efficiency in AUTOSAR~AP.

\vspace{-3mm}
\subsection{Ease of Setting up SOME/IP Conversions}

AUTOSAR~AP uses many IDs, such as ServiceID and InstanceID, which need to be set in the bridge converter.
Just by entering the IDs, a configuration file is automatically generated.
By using the proposed FIDL maker, users can easily create a configuration file and use the bridge converter. 
Table~\ref{tab:input} summarizes the number of lines in the four configuration files and the number of fields to be entered by the user when applied to custom messages in ROS~2 or Autoware and used with the bridge converter.

\vspace{-3mm}
\section{Related Work}
\label{chap:related_work}

Much work have been conducted on AUTOSAR~AP and ROS~2, SOME/IP, and DDS, and other communication protocols. 
This section presents work related to AUTOSAR~AP and ROS 2, SOME/IP and DDS, which are strongly related to this study.
The results of the comparisons are shown in Table~\ref{tab:related_work}.

While the use of SOA is commonplace in autonomous vehicle research in terms of interoperability and scalability, the study by Mehmet et al.~\cite{SOC_in_FAN} introduces a dynamic service-oriented software architecture that provides flexibility in the development of autonomous vehicles. 
The study by Alexandru et al.~\cite{SOA_for_HAV} adds Quality-of-Service (QoS) requirements to SOME/IP for SOA implementation and proposes a framework more suitable for autonomous vehicles.
Furthermore, the study by Marc et al.~\cite{SOA_Orchestration} proposes a resilient architecture with dynamic orchestrators to cope with and manage the steadily increasing amount of software installed in the vehicle.

AUTOSAR~AP is being standardized for scenarios such as highly autonomous vehicles and Car-2-X, according to Simon et al.~\cite{AUTOSAR_AV}, with the goal of smoothly connecting non-AUTOSAR systems while maintaining essential features for autonomous driving systems such as safety, determinism, and real-time functionality.
The study by Jacqueline et al.~\cite{comparison} presents an overall picture of AUTOSAR~AP and ROS~2 and compares the two in terms of how well ROS~2 satisfies the functionalities provided by AUTOSAR~AP. 
Additionally, the study assesses how appropriate ROS~2 is for autonomous vehicles.
The study shows that ROS~2 fully satisfies some of the features of AUTOSAR~AP, but not all of them.

AUTOSAR Adaptive Platform (AP) is currently under standardization to enable the integration of non-AUTOSAR systems. 
As part of this effort, AUTOSAR~AP provides support for various communication protocols beyond SOME/IP, including DDS and Open Platform Communications Unified Architecture (OPC UA). 
However, this support is not yet comprehensive.
The standard also supports protocols other than SOME/IP, such as DDS and Open Platform Communications Unified Architecture (OPC~UA).
The study by Anna et al.~\cite{AUTOSAR_AP_OPC_UA} proposes a new concept of communication via Time-Sensitive Networking (TSN) using OPC~UA in AUTOSAR~AP.
The study by Alexandru et al.~\cite{DDS_OPC_UA} compares DDS and OPC UA and proposes a gateway that enables communication between the two protocols.
The study by Alexandru et al.~\cite{SOMEIP_OPC_UA} also compares SOME/IP and OPC~UA similarly and proposes a gateway that enables communication between the two protocols.
Note that this study is implemented using VSOMEIP.

The study by Alexandru et al.~\cite{gateway} focused on SOME/IP, DDS, and eCAL, comparing their advantages, disadvantages, and compatibility.
They implemented a multi-protocol gateway that allows communication between the three different protocols.
This multi-protocol gateway focuses mainly on communication between AUTOSAR~Classic and AUTOSAR~AP.
This study concentrates on DDS and SOME/IP communication, with a distinct emphasis on facilitating communication between two diverse platforms, AUTOSAR~AP and ROS~2 Adaptive Platform (AP). 
Research on AUTOSAR~AP is still scarce but gradually increasing, and the work of Dongwon et al.~\cite{AUTOSAR_AP_ROS_2} focuses on ROS~2 and AUTOSAR~AP as well as this study. 
The communication between ROS~2 and Adaptive-AUTOSAR~\cite{Adaptive-AUTOSAR}, an open source platform that partially meets the AUTOSAR standard, is verified. 
However, Adaptive-AUTOSAR is not software used in the actual automotive industry and its SOME/IP implementation differs from the AUTOSAR standard. 
Therefore, the triangles are marked in Table~\ref{tab:related_work}.
This study uses the software used in the actual automotive industry, and SOME/IP is also compliant with the AUTOSAR~AP standard.

\section{Conclusion}
\label{chap:conclusion}

This paper proposed a DDS and SOME/IP collaboration framework. 
The proposed collaboration framework bridges the gap between DDS and SOME/IP and enables communication AUTOSAR~AP and ROS~2.
The result is a mix of AUTOSAR~AP and ROS2 in the development of autonomous vehicles. 
The proposed bridge converter is implemented using CommonAPI and vsomeip. 
The bridge converter is implemented as a node of ROS~2, and type conversion and SOME/IP-SD functionalities are added to enable conversion between DDS and SOME/IP. 
The results showed that the time required for conversion is small and that the proposed bridge converter can quickly convert even large data. 
The proposed framework makes useful ROS~2 development support tools such as Rviz and ROSbag, which are not yet available for AUTOSAR AP, available for AUTOSAR AP.
In addition, communication with ROS~2-based software such as Autoware is also possible.
Finally, we implemented FIDL automatic generator to reflect AUTOSAR~AP settings easily in the bridge converter. 
FIDL automatic generator automatically generates FIDL and FDEPL files for this purpose. In addition, by loading an msg file that defines an original custom message, FIDL files describing the conversion specifications can be automatically generated. The use of FIDL automatic generator facilitates the use of the proposed collaboration framework.  

Future challenges include the current support for a limited number of message types and the need to accommodate complex systems requiring multiple message types. 
The proposed bridge converter will evolve by supporting a wider variety of message types and by automating their integration.
QEMU was employed in the AUTOSAR~AP environment for functionality verification during the evaluation. 
However, considering the collaboration between ROS~2 and AP at the control level in an automobile, both must operate on the actual device.
FIDL automatic generator can read the ARXML files used in AUTOSAR~AP and automatically generate the files from them, and we are considering improving it for greater convenience.

\bibliographystyle{IEEEtran} 
\bibliography{refs} 

\end{document}